\documentclass[sigconf]{acmart}

\renewcommand\footnotetextcopyrightpermission[1]{} 
\usepackage{booktabs} 
\usepackage{subcaption}

\usepackage[english]{babel}
\usepackage[utf8x]{inputenc}
\usepackage[T1]{fontenc}
\usepackage{comment} 

\usepackage{amsmath}
\usepackage{graphicx}
\usepackage{xcolor}
\usepackage[colorinlistoftodos]{todonotes}

\setcopyright{none}

\begin{document}

\title[Illuminating the Space of Beatable Lode Runner Levels Produced By Various GANs]{Illuminating the Space of Beatable Lode Runner Levels Produced By Various Generative Adversarial Networks}

\author{Kirby Steckel}
\affiliation{%
  \institution{Southwestern University}
  \streetaddress{1001 E. University Ave}
  \city{Georgetown} 
  \state{Texas, USA} 
  \postcode{78626}
}
\email{steckelk@southwestern.edu}

\author{Jacob Schrum}
\orcid{0000-0002-7315-0515}
\affiliation{%
  \institution{Southwestern University}
  \streetaddress{1001 E. University Ave}
  \city{Georgetown} 
  \state{Texas, USA} 
  \postcode{78626}
}
\email{schrum2@southwestern.edu}

\begin{abstract}
Generative Adversarial Networks (GANs) are capable of generating convincing imitations of elements from a training set, but the distribution of elements in the training set affects to difficulty of properly training the GAN and the quality of the outputs it produces. This paper looks at six different GANs trained on different subsets of data from the game Lode Runner. The quality diversity algorithm MAP-Elites was used to explore the set of quality levels that could be produced by each GAN, where quality was defined as being beatable and having the longest solution path possible. 
Interestingly, a GAN trained on only 20 levels generated the largest set of diverse beatable levels while a GAN trained on 150 levels generated the smallest set of diverse beatable levels, thus challenging the notion that more is always better when training GANs.
\end{abstract}

%
%

\ccsdesc[500]{Computing methodologies~Neural networks}
\ccsdesc[500]{Computing methodologies~Generative and developmental approaches}
\ccsdesc[500]{Computing methodologies~Learning latent representations}
\ccsdesc[500]{Computing methodologies~Genetic algorithms}

\keywords{Lode Runner, Generative Adversarial Networks, Quality Diversity}

\maketitle

\section{Introduction}

Generative Adversarial Networks (GANs \cite{goodfellow2014generative}) are a type of neural network trained in an unsupervised way to produce imitations of elements in a training set. GANs take a latent vector as input and generate convincing fakes based on their training set. The size and composition of the training set influences the quality and diversity of the results from the GAN. Previous work \cite{karras:iclr2018,bontrager2017deepmasterprint} demonstrates that GANs can generate convincing fake results that are indistinguishable from the original training data. GANs have also been applied to video games as a means of Procedural Content Generation (PCG) \cite{volz:gecco2018,giacomello:cog19,torrado:cog20,gutierrez2020zeldagan}.
PCG \cite{shaker2016procedural} is the automatic algorithmic creation of game content, e.g.\ rules, enemies, items, and levels. Use of GANs for PCG is one example of PCG via Machine Learning (PCGML).




PCGML depends on training data, and usually more data is better. However, this paper explores the affects of training with  different subsets of data from the
original 150 levels of the game Lode Runner. Lode Runner is a platform game in which the player runs, climbs, and falls to collect treasure while avoiding enemies. For levels to be beatable, ladders, ropes, and platforms must be organized so that each treasure is reachable. However, not all outputs of a GAN trained on Lode Runner data represent beatable levels. This paper focuses on the challenge of producing beatable Lode Runner levels using GANs, and shows how the size and composition of the training set influences the results. Specifically, the quality diversity algorithm MAP-Elites \cite{mouret:arxiv15} is used to illuminate the search spaces induced by several GANs, to determine the diversity they can produce, and see how many of those levels are actually beatable.



Out of the original 150 levels in the game, training sets of the first 5, 20, 50, 100, and 150 levels were created, as well as a group of 13 levels that are similar in that their layouts depict words and letters.
Contrary to expectations, the GAN trained on all 150 Lode Runner levels did not produce many beatable levels, and was middling in terms of the diversity of the levels produced. Although the diversity of the full training set was larger, the GAN seemed more prone to mode collapse \cite{thanhtung2020catastrophic} as a result. GANs trained on smaller numbers of levels produced both greater diversity and more beatable levels, though the number of diverse and beatable levels drops off again as the size of the training set gets too small. These results demonstrate that smart but limited use of available training data may sometimes lead to better results than simply throwing more data at a problem.



\section{Related Work}
\label{sec:related}



Procedural Content Generation (PCG) is an algorithmic way of generating game content \cite{shaker2016procedural}. The types of content that can be generated are game rules, levels, maps, game items, and more. The use of PCG allows for the developer to automate these processes and reduce their workload, instead of creating all of this game content manually. PCG via Machine Learning (PCGML) uses ML techniques to train PCG systems to create such content.






The ML tool used in this paper is the Generative Adversarial Network (GAN \cite{goodfellow2014generative}). A GAN is a type of neural network that is trained in an unsupervised way to generate convincing fakes based on the training set. GAN training involves two neural networks: a generator (the GAN) and a discriminator. The generator takes real-valued latent inputs, and produces outputs whose size and shape match training samples. The discriminator can either take a real training sample as input, or a fake output from the generator. It produces an output indicating whether it thinks the input is real or fake. Both networks are trained in tandem: the discriminator tries to distinguish between fake and real inputs, and the generator tries to fool the discriminator. Training should stop after
the discriminator can no longer distinguish between real levels from the training set and fake levels from the generator.


GANs have been gaining popularity as a PCGML method for games. The break through work in this domain was the use of a GAN to generate Mario levels \cite{volz:gecco2018}. Since then, the approach has been applied to many other games, such as Doom \cite{giacomello:cog19} and Zelda \cite{gutierrez2020zeldagan}.  

The basic idea of generating levels with a GAN has also been combined with other approaches and enhanced in various ways. The Zelda work mentioned above \cite{gutierrez2020zeldagan} combined GANs with a graph grammar to determine dungeon layouts after the GAN generated rooms. Work in the GVG-AI \cite{gvgai:TCIAIG2016} version of Zelda used Conditional GANs and a bootstrapping method to handle limited data \cite{torrado:cog20}. Another bootstrapping method was applied to an educational puzzle game \cite{park:cog19}. These bootstrapping techniques create more data from the existing data. Both approaches reincorporate certain GAN outputs back into the training set to improve performance. In contrast, a recent approach called TOAD-GAN \cite{awiszus:aiide2020} forgoes the need to bootstrap by generating complete cohesive levels from a single level example. An alternative approach to generating complete cohesive levels is to combine GANs with Compositional Pattern Producing Networks (CPPNs \cite{stanley:gpem2007}), as done recently in the CPPN2GAN approach applied to Mario and Zelda \cite{schrum:gecco2020cppn2gan}.
Another interesting development is the use of one branched GAN to output levels for several different GVG-AI games from a shared latent space \cite{kumaran:aiide2020}.

Other recent works have focused on different ways of searching the latent space induced by trained GANs, such as interactive evolution \cite{schrum:gecco2020interactive}. Another interesting search approach is the quality diversity algorithm MAP-Elites \cite{mouret:arxiv15}, which collects diverse solutions to problems that are also of high quality within their particular niches. MAP-Elites was used in the 
CPPN2GAN \cite{schrum:gecco2020cppn2gan} paper mentioned above, and a specialized variant of MAP-Elites was used in other recent work applied to Mario \cite{fontaine2020illuminating}.



Although previous GAN-based approaches to PCGML are most relevant to this paper, there is another previous PCGML approach to Lode Runner that is worth mentioning. Specifically, Thakkar et al.~\cite{thakkar:cog2019} used standard and variational autoencoders to generate Lode Runner levels. They also evolve new levels by passing existing levels through the autoencoder and mutating the latent representation. There are some similarities to our approach, but this paper differs in the following ways: 1) we evolve latent vectors directly rather than input levels, 2) we use GANs instead of autoencoders, and 3) we use the quality diversity algorithm MAP-Elites instead of pure objective-based evolution. The details are in Section \ref{sec:approach}, but first we provide some relevant details on Lode Runner.



\section{Lode Runner}
\label{sec:loderunner}

Lode Runner was released in 1983 by Broderbund on various home computer systems. The data used in this paper is from the Atari version, which was also used in the paper that generated Lode Runner levels using an autoencoder \cite{thakkar:cog2019}. The game features 150 levels, which are available as part of the Video Game Level Corpus (VGLC \cite{summerville:vglc2016}).
The VGLC stores levels in text files: each tile type from the game is represented by a particular character (Table~\ref{table:tiles}).



Lode Runner is a game that requires the player to traverse platforms and ropes, climb ladders, and fall off cliffs, all while avoiding enemies. The player has a digging tool that can break some ground tiles (there are both breakable and unbreakable tiles), thus creating pit traps for enemies, or new paths for the player to fall down through. The goal of the game is to collect all of the treasure in each level while avoiding enemies. The player cannot  jump, but can catch a rope while falling and fall off the sides of cliffs and ladders.


\begin{table}[t!]
\caption{\label{table:tiles}Tile Types Used in Lode Runner.}
{\small The VGLC character code and integer JSON code for each tile type is shown.
The integer codes are used in the training data derived from VGLC.}

\centering
\begin{tabular}{|c|c|c|c|}
\hline
Tile type & VGLC & Int & Tile \\
\hline
Empty space & \texttt . & 0 & \includegraphics[width=.05\columnwidth]{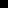} \\
\hline
Gold/treasure & \texttt G & 1 & \includegraphics[width=.05\columnwidth]{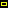} \\
\hline
Enemy & \texttt E & 2 & \includegraphics[width=.05\columnwidth]{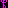} \\
\hline
Diggable ground & \texttt b & 3 & \includegraphics[width=.05\columnwidth]{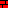} \\
\hline
Ladder & \texttt \# & 4 & \includegraphics[width=.05\columnwidth]{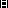} \\
\hline
Rope & \texttt - & 5 & \includegraphics[width=.05\columnwidth]{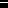} \\
\hline
Solid ground & \texttt B & 6 & \includegraphics[width=.05\columnwidth]{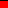}\\
\hline
\end{tabular}
\end{table}

Enemies have the same abilities as the player, with the exception of the digging tool. They can climb ladders, use ropes, and collect treasure. When an enemy collects treasure the only way to get that treasure is to trap that enemy in a pit, causing them to drop the treasure. To beat a level, the player often has to traverse parts of the level multiple times and fall from different locations in order to reach all of the treasure. However, some levels are unbeatable without modelling enemy behavior, because enemies spawn next to treasure that is unreachable for the player. The enemies need to bring the treasure to the player. Some levels are also incorrectly modelled in the VGLC data, because there are insubstantial tiles that the player can fall through. Unfortunately, VGLC just treats these tiles as standard solid tiles. 

Despite these complications, GANs can be trained to produce beatable Lode Runner levels, as discussed in the following section.

\section{Approach}
\label{sec:approach}

The challenge addressed in this paper is the generation of quality beatable Lode Runner levels. First, data from the original game is collected to train several GANs. Latent vectors are evolved for each GAN using MAP-Elites, which collects a diversity of quality levels. MAP-Elites measures the quality of each level using A* search, which determines which levels have the longest solution paths and are therefore the best.


\subsection{Training Data}
First, the VGLC level representations \cite{summerville:vglc2016} are converted into JSON according to Table \ref{table:tiles}, resulting in training sets for the GANs. Each Lode Runner level has the same fixed size, so each level is a single training sample for a GAN.




Training on all 150 available levels did not produce the best results, so different GANs were trained on several subsets of the data. The details on the specific training sets used in our experiments are described in Section \ref{sub:trainingsets}.




\subsection{GAN Training}

\begin{figure}[t]
\centering
\includegraphics[width=\columnwidth]{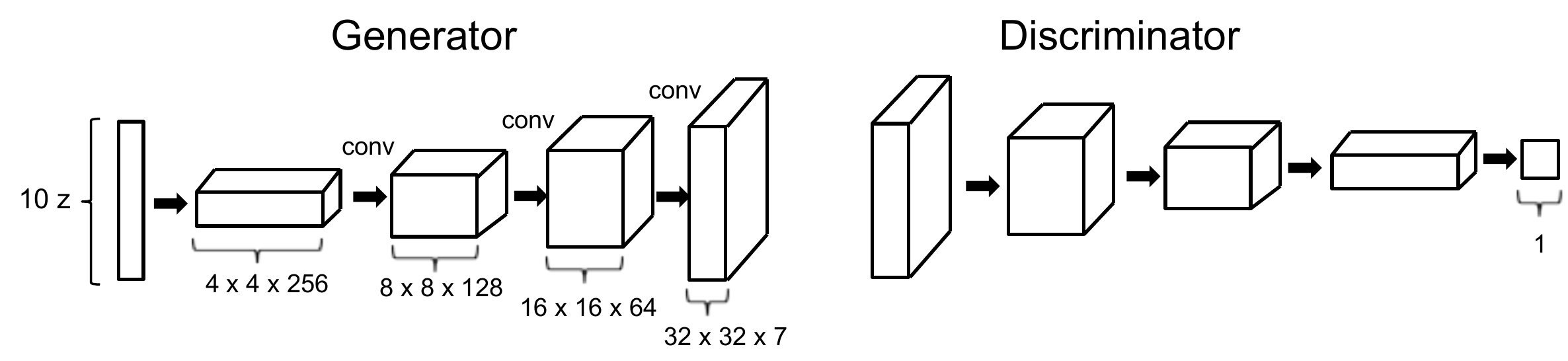}
\caption{GAN architecture.}
\label{fig:architecture}
\end{figure}

The GAN model is trained using the same code\footnote{\url{https://github.com/schrum2/MM-NEAT}} from several recent studies on video game level generation \cite{volz:gecco2018,gutierrez2020zeldagan,schrum:gecco2020cppn2gan,schrum:gecco2020interactive}. The specific model is a Deep Convolutional GAN that is a variant of the Wasserstein GAN \cite{arjovsky2017wasserstein}. The only changes from some previous models are: 1) the latent input size of the generator is now 10, and 2) the depths of both the generator output and the discriminator input are 7 to accommodate the 7 possible tile types in Lode Runner. The architecture is shown in Figure \ref{fig:architecture}

When training, the discriminator accepts inputs from both the training set and the generator. The training set contains 2D representations of levels using the Int codes in Table~\ref{table:tiles}. For input into the discriminator, the 2D levels are expanded into 3D volumes by turning each Int code into a one-hot vector across 7 channels, one for each tile type. The generator outputs the same 3D shape, but for each tile position the maximum output becomes 1 and the rest become 0, resulting in the
needed one-hot encoded format. The 2D representation\footnote{The discriminator input and generator output are $32 \times 32$ for compatibility with other video games in past research. Unused inputs are all 0.} of each level is $32 \times 22$.
 

Each input to the generator is a latent vector of length 10 with values bound to $[-1, 1]$.
The generator represents the learned genotype-to-phenotype mapping once training has been completed. 



\subsection{MAP-Elites}

The trained GANs are used as indirect genotype-to-phenotype mappings for the evolutionary quality diversity algorithm MAP-Elites, which stands for Multidimensional Archive of Phenotypic Elites \cite{mouret:arxiv15}. MAP-Elites maintains an archive of individuals who are the best possible representatives (highest fitness) of their niches, where niches are defined by a binning scheme that discretizes the space of evolved phenotypes along some number of predefined dimensions (see Section \ref{sub:mebins}).

MAP-Elites starts by generating an initial population of 100 random individuals. Each genotype is a real-valued latent vector of length 10. These individuals are placed into bins based on the attributes of the levels that are generated after passing the latent vector through a pre-trained GAN. Each bin only holds one individual, so individuals with higher fitness replace less fit individuals. Once the initial population is generated, solutions are randomly sampled uniformly from the bins and undergo crossover and/or mutation to generate new individuals. These newly created individuals also replace less fit individuals as appropriate, or end up occupying new bins, filling out the range of possible designs. Evolution continues until a fixed number of individuals have been generated.




\subsection{A* Model}
\label{sub:a*model}

The  A* search algorithm is used to determine if a level is beatable and measures the length of the solution path. 
We do not have an actual simulator, so A* uses a simplified model based on the movements that the player can make in the game. 


Player spawn points were not included in the GAN training data because they are so sparse, and there can only be one per level. Thus, the GAN output does not contain a spawn point. Instead, the spawn point is a pseudo-random empty tile in the level. However, to assure that the location of the spawn point actually depends deterministically on the genotype, the pseudo-random generator uses the first of the 10 latent variables in the genome as a seed.



From the spawn point, the player must collect all of the treasure in the level to beat it. The heuristic used by A* is the Manhattan distance to the farthest treasure. Using an exact A* model would have been computationally expensive, so there are several simplifying assumptions in the model. Digging is simplified: players are allowed to move straight down through diggable ground, as long as there is a ground tile to stand on to the left or right of the dig point. The cost for moving down through diggable ground is 4, whereas normal movements have a cost of 1.


Additionally, there are no enemies in our model, both due to the computational cost and because it is unclear what the deterministic policy of the enemy agents would be.
Therefore, some of the original game levels cannot be beaten: as mentioned previously, some levels require the enemy to bring the treasure to the player. 


Because of the high cost of evolving levels, a budget of 100,000 states was imposed on every call to A* search. Such a large number of expanded states usually means that a level is unbeatable, but it is possible that legitimately beatable levels requiring the expansion of more than 100,000 states could be mistaken for unbeatable levels.

Unbeatable levels are assigned a solution path length of -1.
Such levels can be differentiated in terms of \emph{connectivity}.
Connectivity is the percentage of the traversable tiles that the A* model is able to reach. Levels with higher connectivity are presumed to be closer to being beatable, since increasing the connectivity will eventually make all treasure reachable.



\section{Experiments}
\label{sec:experiments}

To determine how the size and composition of the GAN training set affects the ability of MAP-Elites to discover diverse beatable levels, several distinct training sets were defined. Dimensions of variation were also defined for MAP-Elites to set up a binning scheme for the archive. A precise notion of fitness was also needed to determine the best individual for each bin. Finally, various other parameter settings are configured before launching experiments.

 
\subsection{GAN Training Sets}
\label{sub:trainingsets}
Each GAN is trained for 20,000
epochs
on its training set\footnote{GANs were initially trained for 10,000 epochs, which led to poor performance for the larger training sets. Training for 20,000 epochs was a slight improvement, though large training sets still had problems, as seen in the results.}. The specific training sets are all subsets of the 150 levels available in VGLC. Different GANs are trained on the first 5, 20, 50, and 100 levels, as well as the collection of all 150 levels. In the results below (Section \ref{sec:results}), the GANs associated with these training sets are referred to as \texttt{On5Levels}, \texttt{On20Levels}, \texttt{On50Levels}, \texttt{On100Levels}, and \texttt{On150Levels} respectively. There is also a GAN, known as \texttt{WordsPresent},
trained on 13 levels that are similar in that each one depicts words made of several letters. Some examples of words present in these levels are LODE RUNNER, RUSH, and THE END. 

\subsection{MAP-Elites Bins}
\label{sub:mebins}

The levels generated during each experiment get placed into MAP-Elites bins based on their characteristics. Once the appropriate bin for a newly generated level is determined, the bin is checked. If it is empty, the new level is placed in the bin regardless of its fitness. If the bin is occupied, then the new individual can only replace the previous occupant if its fitness is higher.

The appropriate bin for each level is determined based on three different properties: number of enemies, number of treasures, and percentage of ground tiles.
The enemy count is broken up into 10 groups: 9 groups of 2 values each, and a final catchall for larger values:
$\{0,1\}, \{2,3\}, \ldots, \{16,17\}, \{18,19,\ldots\}$. The treasures are broken up into 9 groups of 5 values each and a 10\textsuperscript{th} group for values greater than 44. Ground bins are broken up into 10\% intervals: $[0,10), [10,20), \ldots, [80,90),[90,100]$. Ground percent is the number of diggable and solid tiles divided by the total number of tiles. 


\begin{figure*}[t]
\centering
\begin{subfigure}{0.08\textwidth} 
    \includegraphics[width=1.0\textwidth]{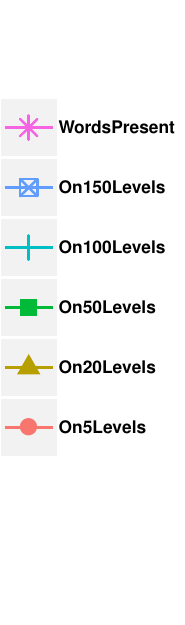}
\end{subfigure}
\begin{subfigure}{0.30\textwidth} 
    \includegraphics[width=1.0\textwidth]{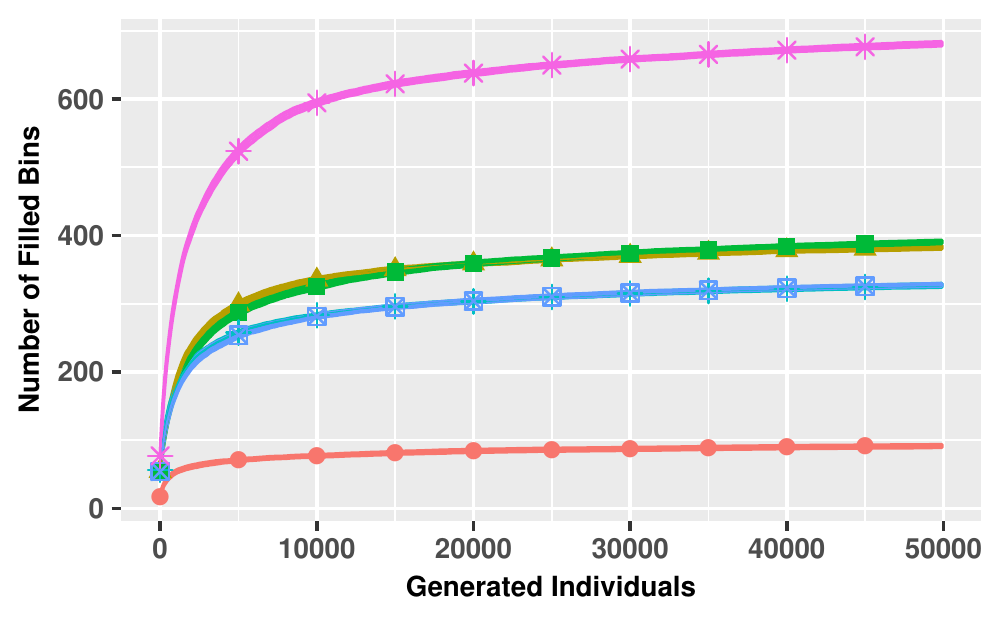}
    \caption{Occupied Bins}
    \label{fig:totalBins}
\end{subfigure}
\begin{subfigure}{0.30\textwidth} 
    \includegraphics[width=1.0\textwidth]{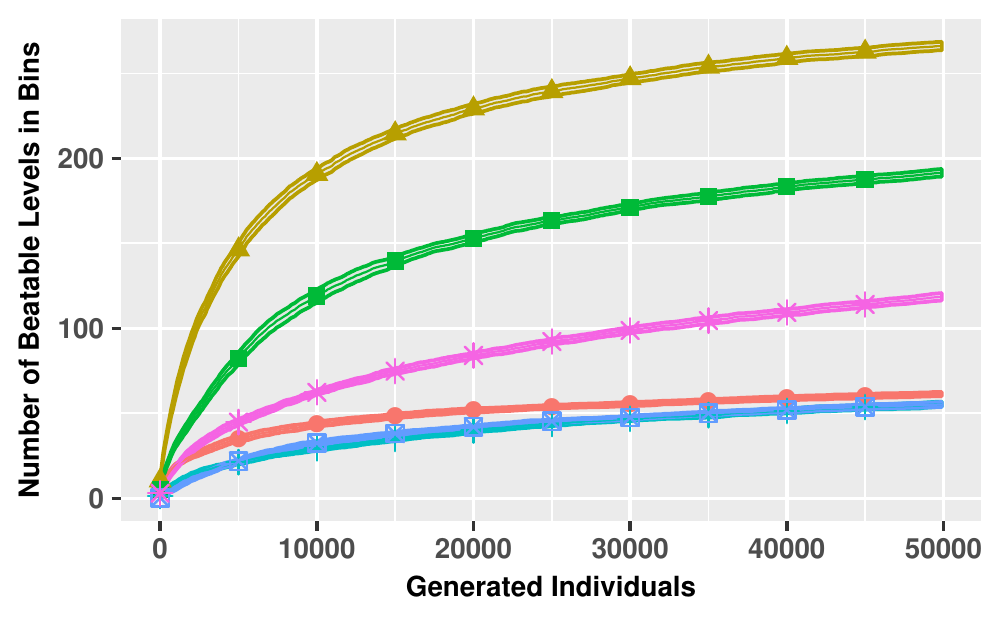}
    \caption{Bins With Beatable Levels}
    \label{fig:beatableBins}
\end{subfigure}
\begin{subfigure}{0.30\textwidth} 
    \includegraphics[width=1.0\textwidth]{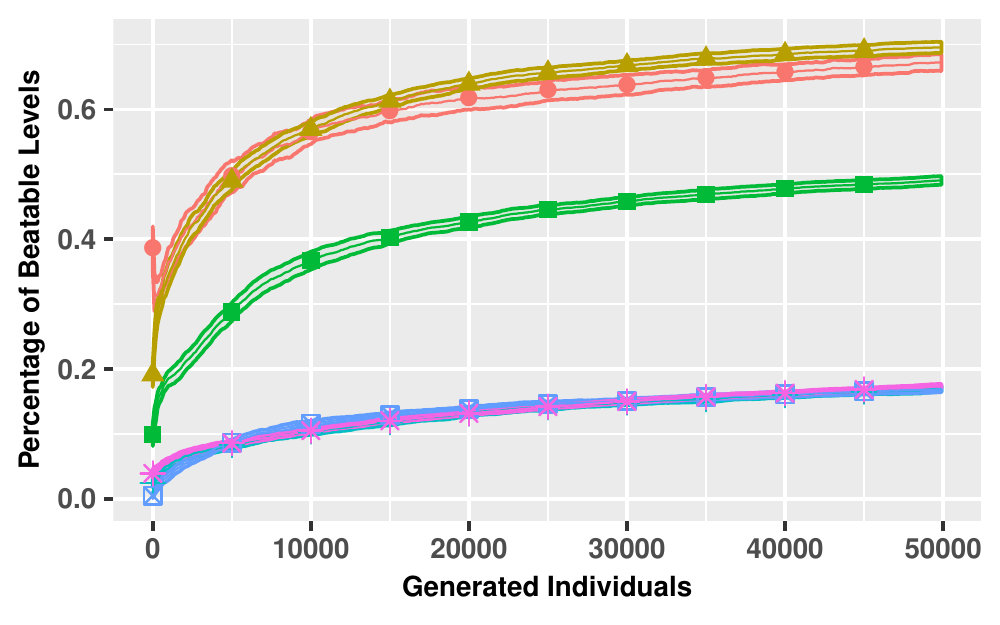}
    \caption{Percent Beatable}
    \label{fig:percentBins}
\end{subfigure}

\caption{Average Filled Bins and Beatable Levels Across 30 Runs of MAP-Elites. \normalfont
(\subref{fig:totalBins}) Average number of
archive bins occupied by a level. 
(\subref{fig:beatableBins}) 
Average number of levels that are actually beatable according to the A* model.
(\subref{fig:percentBins}) 
Percentage of levels that are beatable.
The 95\% confidence intervals are also shown for each line, but are very thin. 
Although the \texttt{WordsPresent} GAN fills the most bins,
the \texttt{On20Levels} GAN produces the most beatable 
levels. \texttt{On5Levels} produces a comparable percentage
of beatable levels to \texttt{On20Levels}, but both the total
number of beatable levels and total number of
occupied bins is very small for \texttt{On5Levels}.}
\label{fig:averageBins}
\end{figure*}

\subsection{Fitness}
\label{sub:fitness}

Fitness is the maximum of the A* solution path length and connectivity, meaning that levels that have a longer solution path are higher quality than those with shorter or no solution path. When there is no solution path, A* search fails and returns a result of~-1. Connectivity is a proportion bound to $[0,1]$, so connectivity only defines the fitness when there is no solution. If a level is beatable, the minimum solution length is 1, which overrides connectivity. 



\subsection{Evolution Parameters}

For each trained GAN, a MAP-Elites archive was evolved 30 times to assure consistent results. Each evolutionary run had an initial population of 100 random individuals, where each genotype was a vector of 10 values in the range $[-1,1]$. Whenever a new individual was needed, there was a $50\%$ chance of single-point crossover, meaning that two random occupied bins were selected to contribute parents that were used to create offspring. Otherwise, one random bin was selected to clone a new offspring vector. However the offspring was created, each real-valued number in the vector had an independent 30\% chance of polynomial mutation \cite{deb1:cs95:polynomial}. Evolution continued until 50,000 individuals were generated.

\section{Results}
\label{sec:results}
Quantitative results describing the bins filled by MAP-Elites are discussed first. Then qualitative results are presented, which focus on the generated levels and their quality. 

\begin{figure*}[p]
\centering
\includegraphics[width=1.0\textwidth]{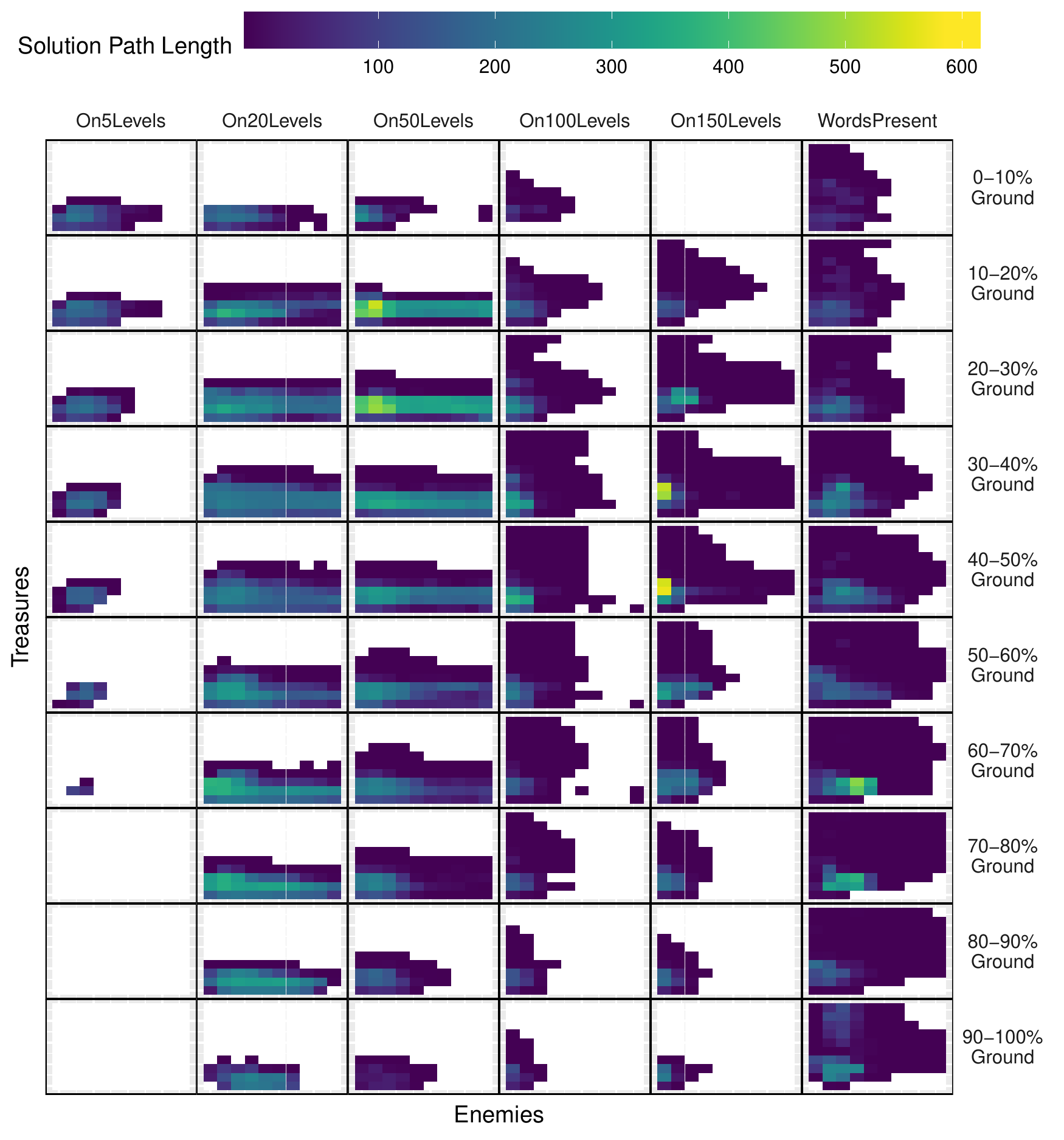}
\caption{Average MAP-Elites Fitness Heat Maps For Each GAN. \normalfont
Each column averages fitness values across MAP Elites runs using a particular
GAN. Rows correspond to the MAP-Elites dimension for percentage of the level
occupied by ground tiles. Each large grid cell is further divided 
into a grid based on the MAP-Elites dimensions of treasure count and enemy
count (treasure count increases from bottom to top, enemy count increases left to right).
White space lacks any evolved genomes, and dark purple corresponds to unbeatable levels.
It is visually clear how \texttt{WordsPresent} best covers the space of possible levels, but
primarily with unbeatable levels. \texttt{On5Levels} has trouble covering the space, especially
for a high percentage of ground coverage. In contrast, the bright colors for
\texttt{On20Levels} and \texttt{On50Levels} indicate a large number of beatable levels.}
\label{fig:averageHeatMaps}
\end{figure*}

\subsection{Quantitative Results} 

Information on how bin occupancy changed over evolution is in Fig.~\ref{fig:averageBins}. Fig.~\ref{fig:totalBins} shows the average number of occupied bins for each GAN during evolution. The \texttt{WordsPresent} GAN had the most bins filled which means that this GAN produced the most variety. The \texttt{On5Levels} GAN had the least variety, which is not surprising given that only five levels were in its training set. 

Fig.~\ref{fig:beatableBins} shows the number of bins with beatable levels for each GAN. The \texttt{On20Levels} GAN had the most beatable levels, followed by the \texttt{On50Levels} GAN. The \texttt{On150Levels} GAN had the lowest number of beatable levels, and the \texttt{On5Levels} GAN was only slightly better. 
The levels generated by the \texttt{On5Levels} GAN were often beatable, but in most cases they were very similar to one of the original five levels \texttt{On5Levels} was trained on. However, it is interesting that \texttt{On5Levels} and \texttt{On150Levels} produce similar numbers of beatable levels, given that the total number of levels from \texttt{On150Levels} that occupy a bin is much larger (Fig.~\ref{fig:totalBins}). 

Fig.~\ref{fig:percentBins} shows the percentage of the levels generated by each GAN that were beatable. The \texttt{On20Levels} GAN ended with the highest percentage of beatable levels followed by the \texttt{On5Levels} GAN, but \texttt{On20Levels} had the highest number of beatable levels as well. These GANs were able to learn the structure of the levels in their training sets and produce beatable levels often. However, the \texttt{On20Levels} GAN was able to generate much more new content than the \texttt{On5Levels} GAN. 
The \texttt{On150Levels} and \texttt{WordsPresent} GANs generated the smallest percentage of beatable levels, which could be due to the complexity of the levels in the training sets. However, even though the percentage of beatable \texttt{WordsPresent} levels is small, the actual number of beatable levels was moderate (Fig.~\ref{fig:beatableBins}), which makes sense given that \texttt{WordsPresent} filled the most bins overall (Fig.~\ref{fig:totalBins}), albeit with mostly unbeatable levels.



Fig.~\ref{fig:averageHeatMaps} is a heat map of average level fitness across all 30 evolution runs for each GAN. 
The grid is split up into 6 columns and 10 rows. Each column corresponds to a single GAN and columns with more white space mean the GAN generates fewer diverse levels according to the MAP-Elites binning scheme described in Section~\ref{sub:mebins}.


Levels with longer solution paths have higher quality and are depicted with brighter colors. Dark purple indicates that the level is not beatable, since this color corresponds to a fitness value less than one. Recall that this means the fitness was defined by its connectivity rather than its A* path length (Section~\ref{sub:fitness}).
Yellow boxes are considered the highest quality levels because they have the longest solution paths.
Each rectangle within each box represents a single level generated by the specified GAN. Rectangles that are closer to the right hand side of their boxes have more enemies and ones closer to the top have more treasures. The right margin shows the ground coverage broken up into 10 intervals of 10\% each, so the first row is when there is 0-10\% ground coverage and so on.

Fig.~\ref{fig:averageHeatMaps} shows that the \texttt{On5Levels}, \texttt{On20Levels}, and \texttt{On50Levels} GANs had low treasure counts. This is because most of the first 50 levels do not have more than 10 treasures. 
The \texttt{On5Levels} GAN does not have high ground coverage because the five levels in its training set have low ground coverage. 
The number of enemies is low for all of the levels from the \texttt{On100Levels} GAN and the levels from the \texttt{On150Levels} GAN that have 50\% ground coverage or more. Interestingly, no levels from the \texttt{On150Levels} GAN have less than 10\% ground coverage.


The heat map is consistent with the results of Fig.~\ref{fig:averageBins} because \texttt{On20Levels} has the highest number of light colored levels and the \texttt{On100Levels}, \texttt{On150Levels}, and \texttt{WordsPresent} GANs have the most dark colored levels. 

\subsection{Qualitative Results}
\label{sub:qualitative}

\begin{figure*}[p]
\centering
\begin{subfigure}{0.24\textwidth} 
    \includegraphics[width=1.0\textwidth]{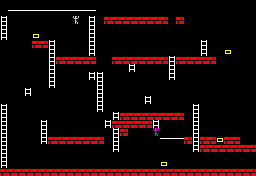}
    \caption{On5Levels Unbeatable}
    \label{fig:On5Levels1}
\end{subfigure}
\begin{subfigure}{0.24\textwidth} 
    \includegraphics[width=1.0\textwidth]{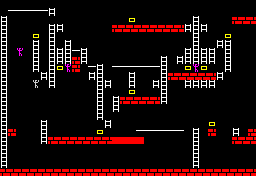}
    \caption{On5Levels Beatable}
    \label{fig:On5Levels2}
\end{subfigure}
\begin{subfigure}{0.24\textwidth} 
    \includegraphics[width=1.0\textwidth]{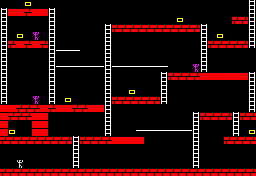}
    \caption{On5Levels Beatable}
    \label{fig:On5Levels3}
\end{subfigure}
\begin{subfigure}{0.24\textwidth} 
    \includegraphics[width=1.0\textwidth]{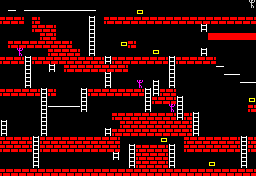}
    \caption{On5Levels Beatable}
    \label{fig:On5Levels4}
\end{subfigure}


\begin{subfigure}{0.24\textwidth} 
    \includegraphics[width=1.0\textwidth]{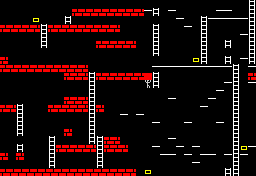}
    \caption{On20Levels Unbeatable}
    \label{fig:On20Levels1}
\end{subfigure}
\begin{subfigure}{0.24\textwidth} 
    \includegraphics[width=1.0\textwidth]{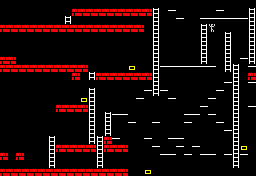}
    \caption{On20Levels Beatable}
    \label{fig:On20Levels2}
\end{subfigure}
\begin{subfigure}{0.24\textwidth} 
    \includegraphics[width=1.0\textwidth]{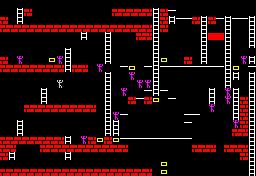}
    \caption{On20Levels Beatable}
    \label{fig:On20Levels3}
\end{subfigure}
\begin{subfigure}{0.24\textwidth} 
    \includegraphics[width=1.0\textwidth]{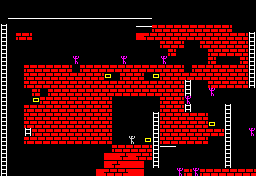}
    \caption{On20Levels Beatable}
    \label{fig:On20Levels4}
\end{subfigure}


\begin{subfigure}{0.24\textwidth} 
    \includegraphics[width=1.0\textwidth]{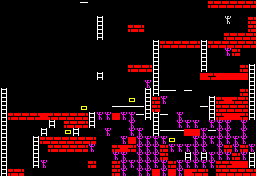}
    \caption{On50Levels Unbeatable}
    \label{fig:On50Levels1}
\end{subfigure}
\begin{subfigure}{0.24\textwidth} 
    \includegraphics[width=1.0\textwidth]{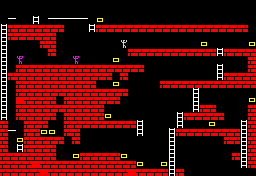}
    \caption{On50Levels Unbeatable}
    \label{fig:On50Levels2}
\end{subfigure}
\begin{subfigure}{0.24\textwidth} 
    \includegraphics[width=1.0\textwidth]{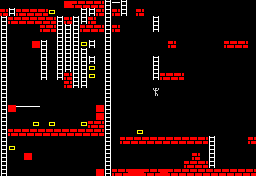}
    \caption{On50Levels Unbeatable}
    \label{fig:On50Levels4}
\end{subfigure}
\begin{subfigure}{0.24\textwidth} 
    \includegraphics[width=1.0\textwidth]{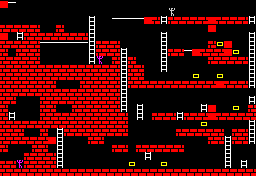}
    \caption{On50Levels Beatable}
    \label{fig:On50Levels3}
\end{subfigure}


\begin{subfigure}{0.24\textwidth} 
    \includegraphics[width=1.0\textwidth]{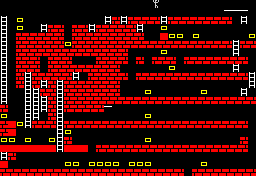}
    \caption{On100Levels Unbeatable}
    \label{fig:On100Levels}
\end{subfigure}
\begin{subfigure}{0.24\textwidth} 
    \includegraphics[width=1.0\textwidth]{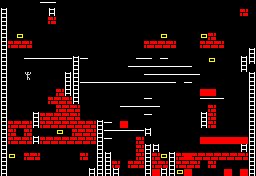}
    \caption{On100Levels Unbeatable}
    \label{fig:On100Levels2}
\end{subfigure}
\begin{subfigure}{0.24\textwidth} 
    \includegraphics[width=1.0\textwidth]{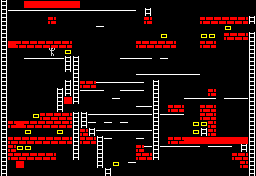}
    \caption{On100Levels Unbeatable}
    \label{fig:On100Levels3}
\end{subfigure}
\begin{subfigure}{0.24\textwidth} 
    \includegraphics[width=1.0\textwidth]{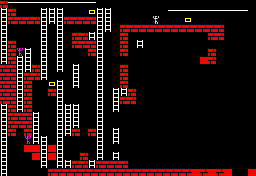}
    \caption{On100Levels Beatable}
    \label{fig:On100Levels4}
\end{subfigure}


\begin{subfigure}{0.24\textwidth} 
    \includegraphics[width=1.0\textwidth]{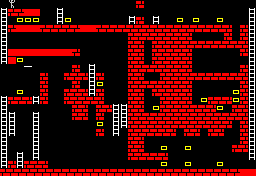}
    \caption{On150Levels Unbeatable}
    \label{fig:On150Levels1}
\end{subfigure}
\begin{subfigure}{0.24\textwidth} 
    \includegraphics[width=1.0\textwidth]{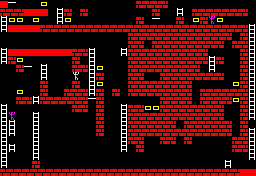}
    \caption{On150Levels Unbeatable}
    \label{fig:On150Levels3}
\end{subfigure}
\begin{subfigure}{0.24\textwidth} 
    \includegraphics[width=1.0\textwidth]{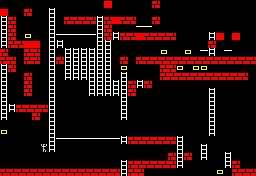}
    \caption{On150Levels Beatable}
    \label{fig:On150Levels2}
\end{subfigure}
\begin{subfigure}{0.24\textwidth} 
    \includegraphics[width=1.0\textwidth]{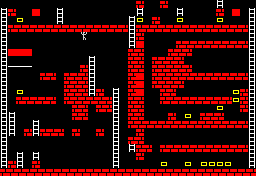}
    \caption{On150Levels Beatable}
    \label{fig:On150Levels4}
\end{subfigure}


\begin{subfigure}{0.24\textwidth} 
    \includegraphics[width=1.0\textwidth]{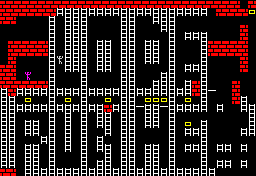}
    \caption{WordsPresent Unbeatable}
    \label{fig:WordsPresent1}
\end{subfigure}
\begin{subfigure}{0.24\textwidth} 
    \includegraphics[width=1.0\textwidth]{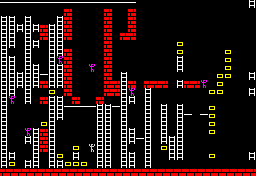}
    \caption{WordsPresent Unbeatable}
    \label{fig:WordsPresent3}
\end{subfigure}
\begin{subfigure}{0.24\textwidth} 
    \includegraphics[width=1.0\textwidth]{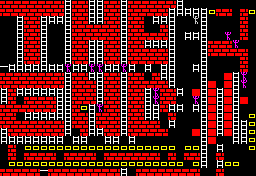}
    \caption{WordsPresent Unbeatable}
    \label{fig:WordsPresent4}
\end{subfigure}
\begin{subfigure}{0.24\textwidth} 
    \includegraphics[width=1.0\textwidth]{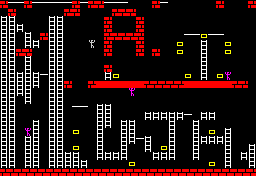}
    \caption{WordsPresent Beatable}
    \label{fig:WordsPresent2}
\end{subfigure}

\caption{Example Generated Levels. \normalfont Each row corresponds to a different GAN. Beatable and unbeatable levels are shown. Several of these levels are referred to in Section~\ref{sub:qualitative}.}
\label{fig:levels}
\end{figure*}

Among the levels generated by the GANs are three common categories: levels that resemble levels from the original game, levels that merge two or more levels from the original game, and levels that are unstructured and chaotic. 

The levels that resemble levels from the original game are more common in the GANs that are trained on fewer levels. Fig.~\ref{fig:On5Levels3} shows a level that is almost an exact copy of the second level from the original Lode Runner game. Since this GAN was only trained on the first five levels, it was easier to encode a mostly complete and accurate representation of all levels in the small training set within the GAN weights. The GANs that are trained on smaller training sets, such as the \texttt{On5Levels} GAN, tend to generate a high number of copied levels. This shows that the GAN is learning the structure of the levels, but is not generating many levels that are new.

However, even GANs with smaller training sets sometimes create original content by composing elements of at least two distinct levels. Levels that merge two or more levels from the original Lode Runner game are present in the output of all GANs. Fig.~\ref{fig:WordsPresent2} is an example of how the GAN has merged existing levels into a new level. In fact, this level has merged features from three different levels. It got the AL on the far left from one level, the A at the top from another level, and the DJA at the bottom from yet another level. Fig.~\ref{fig:On5Levels2} is another example of levels being merged together in the output generated by a GAN. This level gets its platforms and ropes from one level and the structures made of ladders from another level. Fig.~\ref{fig:On5Levels3}, which is a copy of a level from the original game, has the same middle three platforms and ropes. 
These merged levels take elements from different levels to create brand new content. 

The GANs trained on larger training sets produce more unstructured and chaotic output, hence the large number of unbeatable levels generated by the \texttt{On100Levels} and \texttt{On150Levels} GANs. Fig.~\ref{fig:On100Levels2} and Fig.~\ref{fig:On100Levels3} are examples that are unstructured, in that there are several blocks, ropes, and ladders that are simply unreachable and thus serve no purpose. These levels are unbeatable because several treasures are placed high on unreachable platforms. In the original levels, there are instances of enemies needing to bring the treasure down to the player, so these GAN outputs may be emulating those levels. 
However, in some cases unstructured levels can still be beaten, such as in Fig.~\ref{fig:On20Levels2}. This level 
has some unreachable and unnecessary ropes, but is beatable because of the placement of treasure. Fig.~\ref{fig:On50Levels1} is odd because there is a large number of enemies in the bottom right corner. Even if the level were beatable according to A* search, which does not account for enemies (see Section \ref{sub:a*model}), an actual player would very likely not be able to beat the level. 


\section{Discussion and Future Work}
\label{sec:discussion}


GANs trained on more levels were not always better. The GAN trained on all 150 levels was not very successful. If you care about diversity, then \texttt{WordsPresent} is the best GAN to choose. If you care about generating beatable levels, then \texttt{On20Levels} is the best. \texttt{On100Levels} and \texttt{On150Levels} perform poorly in most respects.

The poor performance of \texttt{On100Levels} and \texttt{On150Levels} is potentially due to mode collapse \cite{thanhtung2020catastrophic}. Mode collapse is when the generator produces a reasonable output or small set of outputs and then sticks with them, prioritizing them over other structures. Essentially, the discriminator gets stuck, and the generator takes advantage by repeating the best outputs over and over. Mode collapse should be less likely given that Wasserstein loss \cite{arjovsky2017wasserstein} was used, but it cannot be denied that outputs from \texttt{On100Levels} and \texttt{On150Levels} do not have as much variety, despite being trained on more levels. It is possible that the levels of Lode Runner simply consist of too many unique levels, making it difficult to capture the variety in a large subset of the levels. However, the \texttt{On20Levels}, \texttt{On50Levels}, and \texttt{WordsPresent} GANs were all better at generating levels with more enemies, even though they were trained on fewer samples.


The GANs generate more beatable levels when the training set size is reduced. We can take inspiration from previous bootstrapping experiments \cite{torrado:cog20,park:cog19} and develop an approach in which the GAN is trained first on a smaller subset of the levels, and later has more training data incorporated. As in those previous experiments, we might even incorporate beatable levels produced by the GAN back into the training set as well. The hope is to start from a point of stability, and see if that stability can be maintained while adding training samples. However, it is unclear if either of these approaches would actually produce better level generators, or simply be a lengthier process that still leads to limited outputs.



However, rather than trying to find a way to use all of the data,
one could explore more ways of using limited subsets in interesting ways.
The training set for each GAN only contained the first levels from the 150 original Lode Runner levels. For example, \texttt{On5Levels} was trained on the first five levels only.
The levels at the start of the game all share similar complexity and difficulty.
Training on data sets comprised of random levels from across the training set could lead to different results. If the training samples are spread out across the 150 levels more, it could lead to more diversity in GAN outputs. However, it is also possible that \texttt{On100Levels} and \texttt{On150Levels} perform poorly because later levels in the game do not make good training samples, and using them in any way could be a bad idea. However, such a conclusion would be too defeatist. An ideal method should be able to recreate levels reminiscent of any levels in the training set, no matter how odd.

Another alternative way to define training sets would be to find groups besides the \texttt{WordsPresent} training set that share some kind of structural similarities.
It is interesting that despite only using levels that share the common feature of containing words, \texttt{WordsPresent} actually filled the largest number of bins, and thus according to MAP-Elites actually produces the most diverse outputs. It is unclear how much this result is due to the size of the training set (would any 13 levels do?) and how much depends on these specific training samples.


The A* model used to determine if levels are beatable is very expensive, even though it simplifies the game dynamics. For this reason, a limited search budget was used. However, this limit may have introduced a cap on how complex levels can be. Removing the cap could lead to better results, if one had enough time to run extremely lengthy experiments, but it could also add extra computational expense with no perceivable benefit.



Creating a different MAP-Elites binning scheme could also yield interesting results. The percentage of ground tiles and the numbers of treasures and enemies were chosen as bin dimensions for this paper, but there are alternatives. Other binning schemes could highlight other features of the levels to lead to different results. In fact, for all results for this study, one must keep in mind the caveat that our concept of diversity is tied to the binning scheme used in MAP-Elites. With a different binning scheme, selection pressures would be different, and could even reveal untapped potential in some of the GANs that seem to perform poorly.





\section{Conclusion}
\label{sec:conclusion}

This paper shows that GANs are capable of generating Lode Runner levels using training data from the levels of the original game. 
GANs trained with small to moderate sized subsets of training data 
produced a greater diversity of levels as well as more beatable levels than GANs
trained with larger data sets. Well-trained GANs learn the structure of the levels in the training set to produce new levels with similar characteristics. Although using as much data as possible seems preferable when training a GAN or other PCGML method, these results indicate that others may want to reconsider how much available training data they use with these methods. 

\begin{acks}

This research was made possible by the donation-funded Summer Collaborative Opportunities and Experiences (SCOPE) program for undergraduate research at Southwestern University.

\end{acks}

\bibliographystyle{ACM-Reference-Format}
\bibliography{LodeRunner} 


\end{document}